# Evaluation Sovereignty in Metadata-Driven Classification: A Multi-Track Framework for Weakly Supervised Information Systems

Raymond Vasquez
Lawrence Livermore National Laboratory, Livermore, CA, USA

*The views expressed in this manuscript are those of the author and do not represent the official position of Lawrence Livermore National Laboratory or the U.S. Department of Energy.*

**Abstract**

Evaluation in machine learning is typically treated as a neutral measurement process. However, in operational information systems, evaluation outcomes are often conditioned by the processes used to generate labels. This paper does not aim to improve classification performance. Instead, it examines the validity of performance measurement under differing label authority regimes. This issue is particularly pronounced in large-scale metadata-driven systems, where labels are incomplete, inconsistent, or weakly supervised.

We introduce evaluation sovereignty, defined as the degree to which performance metrics are independent of label authority and supervision regime, and propose a multi-track evaluation framework that systematically varies training and evaluation label sources. Using hierarchical multi-label classification on large-scale scientific metadata, we show that models exhibiting strong performance under operational ("silver") evaluation degrade substantially under independent ("gold") evaluation, particularly for fine-grained classification (e.g., Micro-F1 decreases from ~0.54 to ~0.03). Notably, ranking-based metrics remain non-trivial, revealing a divergence between latent model signal and classification validity.

These findings demonstrate that commonly reported metrics may reflect alignment with labeling processes rather than true predictive capability. We therefore reframe evaluation validity as a system-level property shaped by label governance and provide a practical methodology for auditing intelligent systems under weak supervision.

## 1. Introduction

In intelligent systems and knowledge engineering, classification models are often embedded within larger decision-support workflows. In these settings, evaluation metrics guide system selection and deployment. Misalignment between evaluation and true predictive validity therefore represents a system-level risk, not merely a modeling issue.

Evaluation in machine learning systems is typically treated as a neutral measurement process. In operational information systems, however, evaluation is conditioned by the processes used to generate labels. This paper is not a study of model performance, but of evaluation behavior under differing label regimes. Most evaluations of ML systems use the established measures of performance such as Precision, Recall, and F1 score. These established measures assume that the labels assigned to data for evaluation purposes are valid representations of reality. However, in most cases that is not true. Labels are usually created through some combination of operational practices, indexing procedures, and/or heuristics which create inconsistent and ambiguous labeling along with significant noise (Frénay & Verleysen, 2014; Ratner et al., 2020). Therefore, the reported model performance may actually be representative of the flaws within the labeling process itself rather than the actual predictive capabilities of the model.

There is a major disconnect between what is assumed at the time of evaluation versus the realities of the environment in which ML systems operate. The training and evaluation data are typically sourced from the same place, i.e. the operational system, so there is a great deal of potential for models to appear to perform well due to alignment with the labeling process used during training and evaluation. This creates a form of circular evaluation, whereby models are tested against assumptions similar to those used during training, a phenomenon related to label noise and weak supervision effects (Frénay & Verleysen, 2014). High levels of performance measured by common metrics do not necessarily translate into reliable predictions or generalized prediction. Rather, they tend to reflect the agreement of the model's output with potentially deficient or incomplete supervisory input.

We call this dependency of evaluation outcome on the prevailing label regime, evaluation sovereignty. While demonstrated here in hierarchical multi-label classification, evaluation sovereignty applies broadly to any system in which training and evaluation labels are derived from dependent or partially aligned processes, including weak supervision, human annotation pipelines, and LLM-assisted labeling. Evaluation sovereignty represents the extent to which a model's performance metric(s) are independent of the regime governing its labels. Practically speaking, it describes how the outcomes of evaluation are influenced by the authority, organizational structure, and independence of the labeling regime(s) utilized for assessing a model. Low levels of evaluation sovereignty lead to models being assessed via labels produced from similar processes that were responsible for creating the model's training data; thus, producing potential circularity and artificially elevated assessments. On the other hand, high levels of evaluation sovereignty imply that assessment will utilize labels either independently verified or from higher authorities; thereby, resulting in a more realistic measure of model capacity.

It is particularly relevant in fields relying heavily upon classification systems and large-scale metadata (e.g. scientific information databases and biomedical indexing systems), since classification outputs facilitate search, discovery, and subject-based routing, and subject labels function as training data as well as the primary mechanisms for accessing/retrieving data. Therefore, application-specific evaluation practices reliant upon operational labeling processes used during training pose risks of entrenching metadata workflow variability. Consequently, reported performance may reflect alignment with indexing practices rather than the ability of the classification system to accurately capture document content. This raises a foundational question: How can one evaluate classification systems when labels are products of dynamically changing governance processes rather than fixed representations of truth?

More generally, it is important to distinguish observation from the assumptions used to interpret it. Even highly precise data can produce incorrect conclusions when interpreted within flawed frameworks. For example, in medical diagnosis datasets, models trained and evaluated on clinician-assigned labels may achieve high accuracy by reproducing diagnostic patterns, yet fail when evaluated against independently verified diagnoses. In such cases, performance reflects consistency with the labeling process rather than true predictive validity.

An analogous issue arises in operational classification systems, where evaluation relies on labels generated by the same processes used for training. The concept of evaluation sovereignty formalizes this distinction by emphasizing the independence of evaluation from training label sources, enabling assessment of whether observed performance reflects underlying model capability or artifacts of the data-generating process.

Labels are typically accepted as authoritative regardless of whether they represent variable human judgments, evolving standards or operational limitations. Prior observations in metadata-driven systems, including large-scale repositories such as OSTI, suggest that models are frequently trained and evaluated utilizing labels developed from identical operational pipelines leading to artificially elevated performance metrics.

To help bridge this existing gap we propose a multi-track evaluation framework designed to operationalize evaluation sovereignty by explicitly distinguishing among supervision regimes and label authorities. Our proposed framework will distinguish between operationally-derived ("silver") labels and independently verified ("gold") labels while defining various evaluation conditions that remove or reduce the influence of potential circularity due to differing supervision qualities. Through our proposed framework, we enable explicit measurement of how the relationship between label authority influences evaluation outcomes.

Using large-scale datasets from both scientific metadata and biomedical domains we apply our multi-track framework to evaluate both traditional machine learning models and transformer-based models. Results show that model performance can significantly change as a function of label authority. For example, models performing highly under conditions employing operational labels consistently degrade substantially when evaluated using independently verified labels. Moreover, rank-based metrics can retain signal even after classification accuracy has declined indicating a disconnect between the inherent capabilities of a model and observed evaluation outcomes.

Our results illustrate that validation of evaluations in real-world systems cannot be viewed as an intrinsic property of models themselves. Instead, it needs to be recognized as a systemic attribute dependent on interactions between models, data, and governance processes. Without consideration of label authority in assessments, commonly reported metrics may represent misleading representations of system reliability.

Therefore, the three main contributions of this paper are: (i) We formalize evaluation sovereignty as a property of evaluation independence, (ii) We show empirically that performance metrics are strongly conditioned by label authority and may not reliably reflect model capability under weak supervision, and (iii) We provide a generalizable audit framework for evaluating intelligent classification systems under weak supervision. The models used in this study are not the primary contribution. They serve as controlled instruments to analyze how evaluation behavior changes under different supervision regimes. This has direct implications for model selection, benchmarking validity, and deployment decisions in intelligent systems.

## 2. Background and Related Work

This section reviews prior work relevant to metadata-driven classification, weak supervision, and evaluation practices in machine learning. We focus on how existing approaches address label noise and model performance, and identify a gap in understanding how label authority and supervision regimes influence evaluation outcomes.

Recent advances in large language models (LLMs) have further highlighted the importance of evaluation design, as model performance can vary significantly depending on prompting strategies, label assumptions, and evaluation protocols. These developments reinforce the need to examine evaluation as a system-level construct rather than a fixed benchmark.

### 2.1 Metadata-Driven Classification in Information Systems

Information systems, especially those based on large-scale scientific and technological archives, rely heavily on metadata to provide search functions, help users discover materials, and perform administrative workflow operations. OSTI (Office of Scientific and Technical Information), U.S. Department of Energy, is an example of such an operation. They create and maintain large amounts of archival material and they classify and route each item based on assigned subject labels. As the number of items stored in OSTI continues to grow, assigning subject labels manually is becoming impractical to continue with manual classification of the archive and thus, automated machine learning classification has become a potential solution for automating classification. In addition, text-based automatic document classification has been extensively researched in information science literature (Sebastiani, 2002).

The majority of systems have classification models that have to rely upon readily available descriptive metadata elements such as title and abstract. Although using the limited amount of metadata elements allows for a scalable implementation, it presents several issues including the limitation of contextual information about the classified document and the variability in which the authors describe the contents. These problems are most apparent when using multi-label hierarchical classification where a document can be categorized into multiple levels of abstraction such as broad categories and fine grain sub-categories. Many solutions to this problem include multi-label classification frameworks (Tsoumakas & Katakis, 2008; Silla & Freitas, 2011).

### 2.2 Weak Supervision and Label Noise

One common issue found in all real world classification systems is the use of imperfectly labeled classifications. In most cases, operational labels are created by either indexers performing indexing processes, heuristics rules or semi-automatic tools creating labels. Since these labels are typically produced automatically they will likely include errors, missing labels or other forms of label noise. Previous studies have demonstrated that imperfect labels can negatively affect the way models are trained and tested and ultimately decrease model performance or produce incorrect results (Frénay & Verleysen, 2014).

Weak supervised learning methods and noise tolerant learning methods (Frénay & Verleysen, 2014) were developed to improve the stability of models being trained with noisy labels. Most previous studies focused on providing a method to increase model robustness instead of studying how noisy labels affect our confidence in the evaluation results provided after testing the model. Therefore, although we know a model was trained under noisy label conditions we still cannot confirm whether the performance metrics we report are valid.

### 2.3 Evaluation Assumptions in Machine Learning

Evaluations conducted in machine learning rely on the assumption that evaluation labels represent accurate representations of the ground truth. Accuracy metrics, such as precision, recall, and F1-scores are assumed to measure model accuracy based on this assumption. Due to carefully labeled data sets used in benchmarks, this assumption holds in many cases.

However, due to the fact that training data and evaluation data are often sourced from the same labeling process in operational systems, there exists a risk of circular evaluation where models are evaluated against the same assumptions made during their training process. Thus, models that demonstrate high performance may indicate consistency with the labeling process rather than actual predictive capability. Despite increasing interests in developing data centric methodologies, little emphasis has been placed on understanding how different label authorities influence evaluation outcomes.

**2.4 Evaluation, Label Authority, and Benchmark Validity**

Evaluation in machine learning is often treated as an objective and neutral process; however, a growing body of research challenges this assumption. Recent studies demonstrate that evaluation outcomes are highly sensitive to the design of benchmarks, the quality of underlying data, and the protocols used to generate and validate labels. In particular, benchmark-based evaluation has been criticized for producing results that do not generalize to real-world conditions, leading to what has been described as "phantom progress," where improvements on benchmark datasets do not necessarily correspond to improvements in real-world performance or practical utility (Koohestani et al., 2026; Alampara et al., 2025). Analyses of widely used benchmarks have identified issues including benchmark saturation, data contamination, flawed test cases, and inadequate maintenance, all of which can inflate performance estimates and obscure underlying model limitations. Correcting such deficiencies has been shown to substantially alter reported model performance, suggesting that benchmark quality itself is a critical determinant of evaluation outcomes rather than a neutral measurement instrument (Koohestani et al., 2026).

A central issue in this critique is the mismatch between benchmark conditions and real-world deployment environments. Many benchmarks rely on static datasets, simplified task formulations, and narrow evaluation metrics that fail to capture the complexity, variability, and contextual dependencies present in operational systems (McIntosh et al., 2026; Ashqar, 2025). Furthermore, benchmarks may inadvertently encourage optimization for benchmark performance rather than genuine reasoning, robust generalization, or real-world applicability, leading to inflated assessments of model effectiveness (McIntosh et al., 2026; Orr & Kang, 2024). These concerns are further compounded by data contamination, where overlap between training and evaluation data can artificially inflate performance metrics and obscure true model behavior, thereby weakening the validity of benchmark-based assessments (McIntosh et al., 2026; Koohestani et al., 2026).

Closely related to benchmark limitations is the role of label quality and dataset construction in shaping evaluation outcomes. Research on label noise and data quality has shown that noisy, inconsistent, or weakly supervised labels can significantly degrade model performance, reduce generalization, and alter model behavior (Gharawi et al., 2022). Labeling errors in evaluation datasets are particularly problematic because they directly compromise the validity of performance assessment, influence model selection, and can lead to incorrect conclusions about model capability (Goswami et al., 2023). In large-scale information systems, where labels are often generated through operational processes, indexing workflows, or heuristic procedures, variability in labeling practices introduces additional uncertainty into evaluation, reinforcing the dependence of performance metrics on the underlying label-generation process rather than solely on model capability.

This challenge is further amplified in weakly supervised learning settings, where ground truth labels may be incomplete, noisy, or entirely unavailable. Recent work has framed evaluation under weak supervision as a partial identification problem, in which standard performance metrics such as accuracy, precision, recall, and F1-score cannot be uniquely determined without strong assumptions about the relationship between weak and true labels (Polo et al., 2024). Alternative approaches based on probabilistic performance bounds and partial identification frameworks highlight the inherent uncertainty in evaluating models trained under weak supervision, suggesting that reported metrics may reflect properties of the supervision regime and label-generation process rather than intrinsic model performance (Polo et al., 2024).

Beyond dataset construction, evaluation reliability is also affected by distribution shift between training and deployment environments. When the statistical properties of data change over time or across contexts, models often exhibit significant degradation in performance, and performance measured under conventional evaluation settings may fail to accurately predict model behavior under real-world deployment conditions (Koh et al., 2021; Yao et al., 2022). This breakdown of the independent and identically distributed (i.i.d.) assumption underscores the limitations of conventional evaluation frameworks and highlights the need for evaluation approaches that account for dynamic, real-world conditions and evolving data distributions.

In multi-label and hierarchical classification systems, additional complexity arises from the choice of evaluation metrics themselves. Ranking-based metrics, such as average precision, capture the relative ordering of predicted labels, while classification-based metrics, such as Micro-F1 and Hamming loss, assess binary correctness. Prior work demonstrates that ranking-based and classification-based metrics can produce substantially different assessments of model performance in multilabel classification, with the relative performance of competing methods depending strongly on the selected evaluation metric (García-Pedrajas et al., 2024). This divergence is especially relevant in weakly supervised environments, where models may retain meaningful ranking signal while failing to produce accurate discrete predictions.

Finally, these limitations have motivated broader efforts to reconceptualize evaluation in terms of trustworthiness, accountability, and system-level reliability. Contemporary frameworks emphasize that evaluation must extend beyond accuracy-based metrics to incorporate dimensions such as reliability, robustness, transparency, fairness, accountability, human oversight, and alignment with organizational governance processes (Siavvas et al., 2026; Seralidou et al., 2025). In this view, evaluation is not solely a property of the model, but emerges from the interaction among models, data, human actors, and governance structures that shape how AI systems are designed, deployed, monitored, and interpreted.

Taken together, this body of work suggests that evaluation outcomes are not intrinsic measures of model capability, but are conditioned by benchmark design, label quality, supervision regime, and deployment context. This perspective motivates the need for frameworks that explicitly account for the dependence of evaluation on label authority and data governance, providing a foundation for more reliable and interpretable assessment of intelligent systems.

### 2.5 Gap: Evaluation Under Varying Label Authority

Although previous works explored label noise, data quality, and model performance, no systematic methodologies exist for analyzing how differing levels of label authority affect model evaluations. In particular, none of the current methodologies clearly differentiate between operational labels used for training models and independent verification labels used for evaluating models.

Given that labels are both operational artifacts and evaluation targets in information systems (Frénay & Verleysen, 2014), this gap provides significant motivation for developing an explicit methodology for distinguishing between operational and evaluative labels. Furthermore, without considering varying degrees of label authority, evaluation results may misrepresent model performance and mask inherent limitations. Additionally, since hierarchical classification often includes a long tail distribution of fine grain labels, evaluation metric sensitivity to incomplete labeling processes make this gap particularly relevant in hierarchical classification scenarios.

While prior work has extensively examined label noise, weak supervision, and data-centric approaches to improving model robustness, these efforts primarily treat label quality as a property of the training data and focus on mitigating its effects on model learning. In contrast, this work addresses a distinct but related problem: the validity of evaluation outcomes under varying label authorities. Specifically, we examine how the relationship between training and evaluation label sources influences the interpretation of model performance. By explicitly separating supervision regimes and introducing the concept of evaluation sovereignty, this study reframes evaluation as a dependent variable conditioned on label independence and authority, rather than as an objective measurement of model capability.

To bridge this gap, this study develops a multi-track evaluation methodology that separately considers supervisory regimes and evaluates models relative to the authority level of their respective label sources. Consequently, it advances a practical method for identifying biases in real-world information system evaluations by treating label governance as a variable to assess model reliability. Unlike benchmark validity studies that focus on dataset design, evaluation sovereignty focuses on the relationship between training-label authority and evaluation-label authority as a determinant of observed performance.

## 3. Evaluation as an Object of Study: Label Authority and Supervision Regimes

In most cases, the use of ground-truth labels during training and evaluation of machine-learning classification systems assumes that these labels are consistently reliable and accurate representations of the true relationships within the data. However, in operational systems, this assumption is often violated due to variations in the generation of labels based upon either an operationally-driven process, a heuristic rule-based process, or an operational practice involving human indexers. As a result, the choice of evaluation metric(s) may relate more to the consistency with the labeling process rather than measuring true predictive validity. This challenge has previously been identified in the literature as being one of several important factors related to both the design of evaluations and the selection of thresholds used to interpret the performance of ML models (Lipton et al., 2014).

To better understand these concerns, we introduce a new framework to evaluate classification systems using various levels of label authority. Within this framework, we separately identify training and evaluation labels so that we can examine the influence of training label conditions on performance measured during the evaluation phase.

### 3.1 Label Authority

Label authority represents the extent to which a particular set of labels accurately captures the actual relationships present in the data. In operational systems, labels are produced primarily for operational efficiency purposes rather than to ensure consistent semantic accuracy. Therefore, there will likely be differences in terms of how well these labels capture the true nature of the data. For example, labels generated through the use of evolving taxonomies, institutional policies/practices, and/or human interpretation, do not always adequately capture all aspects of the true structural nature of the data.

Based upon our conceptualization of label authority, we categorize all labels into two types:

- **Silver/Operational Labels**: Labels created through existing operational systems or workflow processes (e.g., metadata indexing, automated tagging, heuristic-based classifications), which can be easily obtained at large-scales; however, they may include errors, inconsistencies, or noise.
- **Gold/Verified Labels**: Labels verified through independent manual reviews, expert annotations, or conformity to specific controlled vocabularies, providing higher accuracy approximations to ground truth. However, because of these verification requirements, they are usually limited in scope or scale.

The distinctions outlined above allow us to gain a clearer picture of how evaluation results vary depending upon the level of authority provided by the labels utilized.

### 3.2 Supervision Regimes and Evaluation Sovereignty

ML models can be trained and evaluated under varying supervision regimes contingent upon whether the labels for training and evaluation were sourced from separate or identical label sources. In many operational contexts, the labels utilized for both training and evaluation are acquired from operational label sources. Although easy to implement, this approach risks validating models against the exact same assumptions inherent within their training data. We formalized this concept below.

Evaluation sovereignty refers to the amount of independence between a given labeling process used for training and the labeling process employed during evaluation. When training and evaluation labels are developed using distinct sources possessing different authorities (i.e., independence exists), it is possible to reduce reliance upon validation performance reflecting properties of a labeling process (versus true predictive ability). Conversely, low evaluation sovereignty occurs when training and evaluation labels originate from the same label sources or processes, thereby creating circular validation where models are validated against their own assumptions about training data. Unlike other concepts (such as sovereign rights), evaluation sovereignty does not exist as a binary attribute. Instead, it exists along a continuous spectrum dependent upon both the level of independence between respective label generating processes and the relative authority each label source contains. A multi-track framework is introduced that provides a means for systematically exploring this range by varying the relationships between training and evaluation label sources (Table 1).

### 3.3 Multi-Track Evaluation Design

To better understand the connection among supervisory regimes and evaluation conditions, we summarize the full multi-track model in terms of the complete combination of all four possible supervision configurations and Track X, an inverted authority condition not tested empirically in this study but important for studying how models that have been trained at high levels of authority interact with models that have been evaluated at lower levels of authority. The focus here is on illustrating that the observed difference in model performance is caused by factors other than model behavior itself, such as the authority level of the labeling systems utilized for evaluation.

**Table 1. Evaluation sovereignty across supervision regimes and label authority conditions**

| Component | Track F (Operational Baseline) | Track R (Authority Gap) | Track P (High-Authority Reference) | Track X (Inverted Authority) |
|---|---|---|---|---|
| **Training Labels** | Silver (Operational) | Silver (Operational) | Gold (Curated) | Gold (Curated) |
| **Evaluation Labels** | Silver (Operational) | Gold (Verified) | Gold (Curated) | Silver (Operational) |
| **Label Source Relationship** | Same source (dependent) | Different sources (independent) | Same source (independently governed) | Different sources (independent) |
| **Evaluation Sovereignty** | **Low** | **High** | **High** | **Low (inverted)** |
| **Interpretation of Performance** | Potentially inflated (circular) | Authority gap revealed | Reliable estimate of capability | Excellence potentially suppressed |
| **Primary Role in Study** | Operational baseline | Authority gap test | High-authority benchmark | Theoretical — not empirically evaluated in this study |

| Component | Track F (Operational Baseline) | Track R (Authority Gap) | Track P (High-Authority Reference) | Track X (Inverted Authority) |
| --- | --- | --- | --- | --- |
| **Dataset Context** | OSTI Metadata | OSTI (Silver → Gold subset) | PubMed / MeSH | Not evaluated in this study |

As demonstrated in Table 1, each configuration represents a different evaluation track; each track includes a unique combination of the source of labels used for training/evaluation:

- **Track F (Operational baseline)**: trained and evaluated using operational labels (silver). Standard operating practices in most operational environments utilize this type of label; however, this approach is vulnerable to circular evaluation effects.
- **Track R (Authority gap evaluation)**: trained using operational labels (silver), evaluated using independently verified labels (gold). Track R evaluates the gap between the performance achieved when models operate based upon operational assumptions vs. The performance when models are evaluated using more authoritative evaluation conditions.
- **Track P (External validation)**: both training and evaluation use external curation-standardized label sets. Provides a baseline of model performance using independently defined/governed taxonomic standards.
- **Track X (Inverted authority)**: trained using gold labels, evaluated against operational silver labels. Inverted authority is expected to result in artificially depressed apparent performance if the model was trained to a higher authority standard, it will likely receive poor scores when evaluated using low-authority labels as they are unable to recognize the distinctions made by the model through gold-label training. Future work should investigate this configuration.

**Interpretation of the Matrix**

The supervision-evaluation matrix provides a systematic way of examining the effect of evaluation conditions on observable performance. Specifically:

- **Silver → Silver (Track F)**: when the evaluation labels are aligned with the training labels (i.e., both silver), the model's overall performance will reflect consistent application of the operational labeling system. However, this can also cause an inflationary bias in the model's performance as there is no disambiguation between the two.
- **Silver → Gold (Track R)**: enforces independence between training and evaluation labels, revealing the extent to which model performance depends on operational labeling assumptions. This condition should be interpreted as an audit of evaluation validity rather than a definitive measure of predictive accuracy.
- **Gold → Gold (Track P)**: serves as a reference point for evaluating model performance under high-authority training. As both training and evaluation use independently curated labels, this provides confidence that model performance reflects actual performance.
- **Gold → Silver (Track X)**: due to the lower authority level of the evaluation labels, model performance may be artificially suppressed. If the model has learned to apply a classification standard that exceeds what is provided by the operational labeling system (e.g., encodings of precision), it would likely appear to perform less well compared to similar models trained under silver-only conditions.

A substantial decrease in performance from Track F to Track R indicates that model predictions are largely dependent upon properties of the operational labeling system rather than semantic correctness.

**Framework Implications**

The multi-track evaluation design expands evaluation from a single metric-based measurement of model performance to provide a systematic methodology for assessing the reliability of classification systems under real-world conditions. Rather than assume that evaluation labels represent true authority, the framework systematically tests this assumption and quantifies its effects.

Additionally, by providing various supervision regimes, the framework allows practitioners to differentiate between models that perform effectively within operational constraints and models that continue to maintain their performance when subjected to more authoritative evaluation conditions. Understanding these distinctions are critical when considering model behavior in information systems where labels are both operational artifacts and evaluation targets.

## 4. Experimental Setup

The experiment setup enabled the investigation of the impact of the different levels of label authority on the performance of the proposed multi-track framework. The experiments were conducted based on a large amount of data that can be found within two main domains of scientific metadata and biomedical literature.

### 4.1 Datasets

We use two primary datasets to represent operational and externally curated labeling environments:

**OSTI Dataset (Scientific Metadata)**:

A large collection of data from scientific and technical records taken from the U.S. Department of Energy's Office of Scientific and Technical Information (OSTI) database. Each record contains several metadata fields including the title, an abstract, and subject labels. Labels that are applied to subject categories are intended to create hierarchical multi-label classification problems. Based on past research, all input to models was limited to metadata fields (*Title* and *Description*) to simulate the limitations present in actual deployments. A total of 1000 OSTI records were randomly selected and manually reviewed and validated by the author through established documentation to define the gold standard evaluation set for Track R. This set was strictly excluded from being used for training or model fitting. No measure of inter-annotator agreement was performed. Accordingly, Track R should be interpreted as an audit condition designed to enforce independence between supervision regimes, rather than as a definitive or externally validated ground-truth benchmark.

**PubMed Dataset (Biomedical Literature)**:

A pre-curated set of articles published in biomedicine with MeSH assigned labels. In addition to providing an environment with independently managed labels it allows for testing the performance of supervised learning methods when there are high levels of authority controlling the labels.

In terms of the differences in how the two datasets have been developed they both provide different viewpoints. The development of the OSTI dataset simulates the real-world operational labeling that occurs in practice whereas the PubMed dataset represents a labeling environment that has been standardized and is maintained outside of an organization's control.

### 4.2 Label Processing and Hierarchical Structure

The datasets were processed to allow hierarchical multi-label classification. All of the labels were split into two layers:

- **Level-1 (L1)**: General Subject Categories
- **Level-2 (L2)**: More Specific Keywords and Descriptors

Labeling inconsistencies were addressed through normalization procedures which normalized variant spellings and synonyms to their canonical label form. This process was done to prevent variations in labeling from artificially inflating or deflating model performance. A long-tail distribution is found within the final label space, especially at Layer 2, as there are many labels that appear very rarely

### 4.3 Models

Classical and Transformer-Based models were used to assure that the effects seen were not based upon the use of a particular Model Architecture.

- **Classical Models**:
  - Term Frequency–Inverse Document Frequency (TF-IDF) representations combined with Logistic Regression classifiers (Pedregosa et al., 2011)
  - One-vs-Rest (OvR) and classifier chain variants (Read et al., 2011)
- **Transformer-Based Models**:
  - Pretrained language models, including BERT-based and related architectures (Devlin et al., 2019; Liu et al., 2019; Beltagy et al., 2019; Beltagy et al., 2020; Gu et al., 2020)
  - Fine-tuned for multi-label classification using metadata inputs

All of the models are trained using an equivalent input representation (*Title* and *Description*/*Abstract*) and evaluated using the same supervision regimes to provide equivalence between evaluation tracks.

### 4.4 Data Splits and Training Protocol

The datasets were divided into training, validation and testing subsets with a fixed split so that each study could be repeated as needed. When possible stratified sampling was also used to maintain the same label distributions among the data subsets. The best combination of model parameters was determined from the validation subset and all performance assessments for the test data will be made without access to it.

Thresholds for multi-class labeling were developed based solely upon performance on the validation subset. Thresholds established this way cannot leak knowledge about test performance back into determining which model parameters were chosen; thus, when comparing models' performances across tracks the difference in their supervised learning environments can be compared independent of the model parameterization chosen. All dataset splits, label encoders, and evaluation configurations were preserved to support reproducibility.

### 4.5 Evaluation Metrics

The model effectiveness was assessed through multi-class classification metrics:

- **Micro-averaged Precision, Recall, and F1-score**: Capture overall performance across all labels, weighted by label frequency
- **Macro-averaged Precision, Recall, and F1-score**: Reflect performance across labels equally, highlighting behavior on rare classes
- **Area Under the Precision–Recall Curve (AUPRC)**: Provides insight into ranking quality across decision thresholds
- **Hamming Loss**: Measures the fraction of incorrectly predicted labels

Performance is measured uniformly throughout all assessment tracks to allow for comparative analysis of performance using varying levels of supervision.

### 4.6 Experimental Design Across Multiple Tracks

This experiment's methodological design is based on the multiple track framework described in Section 3. For each model, performance is evaluated under the following conditions:

- **Track F (Baseline-operational)**: Training and evaluation are conducted using operational (silver) labels derived from the OSTI dataset.
- **Track R (Evaluation authority gap)**: in this category, models that were trained using operational labels will have their accuracy measured by comparing them to an independent gold standard. Thus, we can measure the differences caused by label authority (i.e., how much better does a model perform when its data has been labeled using "gold" instead of "silver").
- **Track P (Validation using externally governed labels)**: in this category, the models will be trained and tested using a dataset of curated labels found in the PubMed database. We use these results as a comparison for measuring performance while being supervised by external governing entities.

Using similar databases for both training and testing, along with identical models and evaluation criteria for the different categories allows us to isolate the effects of label authority on model performance.

## 5. Results

We measured model performance under all aspects of the multi-track framework so we could determine what influence(s) label authority has upon observed outcomes. We used identical metrics and model settings across all track levels and provide results for both Level-1 (L1) and Level-2 (L2) label space.

### 5.1 Model Performance Under Operational Baseline (Track F)

Using an operational baseline (Track F) where both the train set and test set contain "silver" labels, our models exhibited strong performance under operational evaluation conditions when applying common evaluation metrics. Transformer-based models significantly outperformed traditional models in terms of micro-averaged F1 and AUPRC. Traditional TF-IDF + Logistic Regression models remained competitive in L1 label space, however their performance was greatly diminished in L2 label space due to the much greater sparseness and long tail distribution of fine grained labels.

In L2, the application of a Top-N ranking constraint (i.e., top 100) helped stabilize the evaluation and prevented the models from producing excessive numbers of poor confidence labels. This method of evaluating models via ranking is also supported by previous studies regarding evaluation methods that involve thresholding and ranking within multi-label classification (Tsoumakas & Katakis, 2008; Lipton et al., 2014).

Importantly, the use of a Top-N constraint alters the interpretation of classification metrics at Level-2. Under this setting, models are evaluated based on their ability to recover relevant labels within a bounded candidate set, effectively framing the task as a constrained ranking problem rather than unconstrained multi-label classification. As a result, Micro-F1 scores reported in Track F reflect performance under a ranking-constrained evaluation regime and are not directly comparable to strict classification outcomes without such constraints.

When evaluated under the Top-N constraints, both traditional and transformer-based models produced moderate to strong performance (Micro-F1 ≈ 0.61–0.64 at Level-1), with the transformer-based models being the best performers in terms of micro-F1 and AUPRC. These results support typical expectations for models trained in this manner, indicating that models have learned identifiable patterns present in the operational labeling structure. Note that the Top-N constraint is not intended to improve classification performance, but to stabilize evaluation under extreme label sparsity by reframing the task as bounded candidate recovery.

### 5.2 Authority Gap: Silver-to-Gold Evaluation (Track R)

Performance for all model types declines significantly under Track R evaluation conditions, where models initially developed with operational ("silver") label data are compared with independently validated ("gold") labels. This decline should be interpreted as an effect of the evaluation regime rather than solely as a limitation of model capability.

L1 performance drops from moderately high Micro-F1 scores under Track F to lower values under Track R (e.g., from approximately 0.64 to 0.37), indicating limited overlap between the operational and validated label sets. At Level-2, classification performance approaches near-zero levels under strict evaluation criteria (Micro-F1 decreases from approximately 0.54 to 0.03). Although models may still produce non-trivial rankings of labels (i.e., ranking metrics remain above baseline), these rankings are generally misaligned when compared to the verified label set.

The magnitude of the observed performance decline at Level-2 must be interpreted in light of both label authority and evaluation protocol differences. In Track F, Level-2 evaluation is conducted under a Top-N ranking constraint, which stabilizes predictions by limiting outputs to the highest-scoring candidate labels. This effectively evaluates the model's ability to rank relevant labels within a bounded set rather than requiring precise classification across the full label space.

In contrast, Track R removes this ranking constraint and evaluates predictions against independently verified gold labels using strict classification criteria. This transition introduces a substantially more stringent evaluation condition in a sparse, high-dimensional label space, where correct classification requires precise alignment with independently verified labels. Consequently, the observed decline (Micro-F1 from approximately 0.54 to 0.03) reflects not only the authority gap between operational and verified labels, but also the shift from ranking-constrained evaluation to unconstrained binary classification. This combined effect highlights that observed performance differences across tracks are both semantic (label authority) and methodological (evaluation protocol).

Although evaluation protocols differ between Track F (Top-N constrained) and Track R (strict classification), the magnitude of performance collapse exceeds what would be expected from protocol differences alone, indicating a dominant effect of label authority.

### 5.3 External Validation Under Curated Labels (Track P)

The consistency and strength demonstrated by models under Track P indicate that models can learn effectively under conditions where label authority is high and consistent. Performance recovers under curated supervision, reaching approximately 0.95 Micro-F1 at Level-1 and 0.83 at Level-2, with similarly strong AUPRC values across both classical and transformer-based models.

These results demonstrate that strong performance is achievable when both training and evaluation are governed by high-authority, consistently curated labels. However, this does not imply that such performance transfers to operational environments, where labeling processes differ substantially in both structure and governance.

Importantly, the contrast between Track P and Track R highlights that the degradation observed in Track R is not solely due to model limitations. Rather, it reflects the effect of mismatched label authority between training and evaluation conditions. When this mismatch is removed, as in Track P, model performance remains strong, indicating that the primary constraint in Track R is the evaluation condition rather than the model's ability to learn.

### 5.4 Ranking vs. Classification Behavior

Beginning with an analysis of ranking-based vs. classification-based evaluations across all track types, we observed very similar trends across both types of metrics. For example, in Track R, although binary classification performance at Level-2 approaches chance levels, the Area Under the Precision–Recall Curve (AUPRC) and Label Ranking Average Precision (LRAP) values remain above baseline.

This is consistent with prior work highlighting the usefulness of ranking-based metrics in multi-label classification (Tsoumakas & Katakis, 2008). The fact that models retain some rank-ordering information concerning labels, even though they do not produce strong predictions under strict evaluation criteria, further supports the utility of such metrics.

As a direct result of these findings, it appears that models trained on weakly labeled data are able to infer partial semantic relationships between items; however, these models fail to translate this relational signal into correct classification decisions when evaluated relative to higher-authority labels.

Thus, caution is required when assessing model performance in weakly labeled environments. In particular, it is important to distinguish between measures of model quality based on ranking and measures of predictive accuracy based on classification. This distinction is especially relevant when interpreting Track F results, where Top-N constraints allow models to satisfy ranking-based performance criteria while lacking the precision required for accurate classification under higher-authority conditions.

### 5.5 Summary of Cross-Track Findings

Taken together, the empirically evaluated tracks reveal three consistent patterns:

- **Inflated Performance Under Operational Evaluation:** Models achieve strong performance when evaluated on the same operational labels used for training, indicating potential circularity.
- **Significant Degradation Under Gold Evaluation:** Performance drops substantially when evaluated against independently verified labels, particularly for fine-grained (Level-2) classification.
- **Recovery Under Curated Supervision:** Models perform strongly when both training and evaluation rely on high-authority, externally governed labels.

Table 2 presents L1 classification performance across all three tracks for representative classical and transformer-based models. The contrast between Track F and Track R performance illustrates the authority gap directly. SciBERT achieves a Micro-F1 of 0.6380 under operational evaluation, while the best-performing model under Track R degrades to 0.3699 under gold evaluation, and models under Track P recover to 0.9495 under curated supervision.

**Table 2. Level-1 classification performance across supervision regimes.**

| Track | Supervision Regime | Model | Micro-F1 | Macro-F1 | Micro-AUPRC |
|---|---|---|---|---|---|
| **F (Operational)** | Silver → Silver | SciBERT | 0.6380 | 0.4269 | 0.6762 |
| **F (Operational)** | Silver → Silver | TF-IDF + LR (HCC) | 0.6095 | 0.4171 | 0.6279 |
| **R (Authority Gap)** | Silver → Gold | TF-IDF + LR (OvR) | 0.3699 | 0.0998 | 0.3099 |
| **R (Authority Gap)** | Silver → Gold | TF-IDF + LR (CC) | 0.3653 | 0.0984 | 0.3111 |
| **P (External Gold)** | Gold → Gold | SciBERT | 0.9495 | 0.4445 | 0.9828 |
| **P (External Gold)** | Gold → Gold | PubMedBERT | 0.9465 | 0.4226 | 0.9800 |

*Track F (Operational): silver-trained, silver-evaluated. Track R (Authority Gap): silver-trained, gold-evaluated. Track P (External Gold): gold-trained, gold-evaluated. Track R demonstrates substantial performance degradation when silver-trained models are evaluated against gold authority labels, despite comparable Track F operational performance.*

Beginning with the table for the L2 data (Table 3), we find the authority gap is most apparent here. With the near complete failure of the Micro-F1 in Track R, going from an operational Micro-F1 of 0.5380 to a gold evaluated Micro-F1 of 0.0273, it is clear that fine-grained label predictions are highly sensitive to evaluation label authority. However, as seen in Section 5.4, the remaining LRAP score values in Track R are significantly higher than the Micro-F1 scores; this reflects the distinction between ranking-based and classification-based behavior.

**Table 3. Level-2 classification performance across supervision regimes.**

| Track | Supervision Regime | Condition | Micro-F1 | Macro-F1 | LRAP |
|---|---|---|---|---|---|
| F (Operational) | Silver → Silver | Baseline ANY_TRIM | 0.5380 | 0.4556 | — |
| F (Operational) | Silver → Silver | +SHAP+Dict STRICT_ONLY | 0.7303 | 0.1659 | — |
| R (Authority Gap) | Silver → Gold | TF-IDF + LR (OvR) | 0.0273 | 0.00043 | 0.5815 |
| R (Authority Gap) | Silver → Gold | TF-IDF + LR (CC) | 0.0157 | 0.00015 | 0.5755 |
| P (External Gold) | Gold → Gold | SciBERT | 0.8112 | 0.1274 | 0.8805 |
| P (External Gold) | Gold → Gold | PubMedBERT | 0.8350 | 0.1448 | 0.9053 |

*Track F (Operational): silver-trained, silver-evaluated. Track F results are computed under Top-N (100) constrained evaluation. Track R (Authority Gap): silver-trained, gold-evaluated. Track P (External Gold): gold-trained, gold-evaluated. Track R collapse at Level-2 from operational Micro-F1 of 0.54 to gold-evaluated Micro-F1 of 0.03 provides the clearest empirical demonstration of the evaluation sovereignty effect. LRAP = Label Ranking Average Precision. Dashes indicate metric not computed for this condition.*

Figure 1 provides a consolidated view of L1 and L2 Micro-F1 performance across supervision regimes:

| Track | Supervision | L1 Micro-F1 | L2 Micro-F1 |
|---|---|---|---|
| F (Operational) | OSTI (Silver) | 0.64 | ~0.54 (Top-100, diagnostic) |
| R (Authority Gap) | Silver → Gold | 0.37 | 0.03 |

| Track | Supervision | L1 Micro-F1 | L2 Micro-F1 |
|---|---|---|---|
| **P (External Gold)** | External Gold (PubMed/MeSH curated labels) | 0.95 | 0.83 |

**Figure 1. Classification Performance Across Supervision Regimes (Tracks F, R, and P).** Performance varies as a function of label authority and evaluation protocol**.** *Note: Track F Level-2 results are derived under Top-N constrained evaluation and are not directly comparable to strict classification results reported for Tracks R and P.*

As shown in Figure 1, Level-1 performance remains relatively stable across supervision regimes, while Level-2 performance varies dramatically, with near-complete collapse under Track R and strong recovery under Track P. This contrast highlights that evaluation outcomes are conditioned by both label authority and evaluation protocol, rather than reflecting intrinsic model capability alone.

These observations show how strongly model effectiveness depends on the authority of the labels used for evaluation. A model may be perceived as being very successful based on its operational assessment and unsuccessful if assessed with an independent validation (gold) label. Therefore, this observation shows how greatly different assessments by different regimes of supervision can influence what is measured.

## 6. Discussion

Importantly, the findings of this study should not be interpreted as a validation of classification performance within the OSTI system. Rather, the results constitute an empirical audit of evaluation regime in intelligent classification systems and label authority, illustrating how evaluation outcomes depend on the relationship between training and evaluation label sources. The observed performance differences across tracks reflect properties of the evaluation conditions rather than intrinsic differences in model capability. For all tested models and data sets, we demonstrate that model performance varies substantially depending on the authority of the labels used for evaluation; i.e., either the "operational" (silver) labels provided during training or the "independently validated" (gold) labels that were created after each test was run. Our research results challenge the idea that well-established measures of classification model performance will provide a consistent and objective view of the performance of such models.

A key observation is the presence of inflated performance under operational evaluation conditions. This effect is consistent with prior findings on label noise and weak supervision (Frénay & Verleysen, 2014). In Track F, where training and evaluation rely on the same operational label source, models achieve strong results across multiple metrics. However, these results are not necessarily indicative of true predictive validity. Instead, they reflect consistency with the labeling system itself. Because both training and evaluation labels originate from the same process, models are effectively optimized to reproduce the patterns embedded in that process, including any biases or inconsistencies it may contain.

This effect becomes evident when moving to Track R, where models trained on operational labels are evaluated against independently verified labels. The substantial drop in performance observed in this setting highlights the authority gap between operational and verified supervision. At Level-2, this gap is particularly pronounced, with binary classification metrics approaching near-zero values despite non-trivial ranking performance. These results indicate that models trained under weak supervision may capture partial signal but fail to produce reliable predictions when evaluated against higher-authority labels. While evaluation protocol differences contribute to observed variation, the dominant effect arises from label authority

The divergence between ranking-based and classification-based metrics provides further insight into model behavior under weak supervision. While models retain some ability to rank relevant labels, as reflected in AUPRC and related measures, they are unable to translate this signal into accurate discrete predictions. This suggests that the underlying representation learned by the model contains useful information, but that the decision boundaries required for reliable classification are misaligned with the true label structure. In practical terms, this means that models may be useful for candidate generation or prioritization tasks, but not for definitive classification without additional validation.

The strong performance observed in Track P confirms that the models themselves are not inherently limited. When both training and evaluation are conducted using high-authority, externally curated labels, models achieve consistent and reliable results. This indicates that the primary constraint in Track R is not model capacity, but rather the mismatch between training and evaluation label authority. The implication is that improving model architecture alone is insufficient to address performance issues in weakly supervised systems; attention must also be directed toward the quality and governance of the labeling process.

These findings support the central premise of the proposed framework: evaluation validity is a system-level property, not a purely model-level characteristic. Model performance depends not only on the algorithm and input data, but also on how labels are generated, curated, and used for evaluation. In systems where labels serve both operational and evaluative roles, this dual function can obscure the true behavior of the model and lead to overconfident interpretations of performance.

From a practical perspective, the multi-track framework provides a structured methodology for diagnosing these issues. By explicitly separating supervision regimes and comparing performance across tracks, practitioners can identify when evaluation results are likely to be inflated and when they reflect more reliable predictive behavior. This approach enables more informed decisions about model deployment, particularly in high-stakes environments where incorrect predictions may have significant consequences.

More broadly, the results highlight the importance of integrating data governance considerations into machine learning system design. Rather than treating labels as static inputs, the framework emphasizes their role as dynamic artifacts shaped by organizational processes. This perspective aligns with emerging work in data-centric artificial intelligence, but extends it by focusing specifically on the implications for evaluation.

Finally, the concept of *evaluation sovereignty* provides a useful lens for understanding these dynamics. By framing evaluation as dependent on the independence of label sources, the framework offers a principled way to assess when performance metrics can be trusted. Systems that lack evaluation sovereignty are more susceptible to circularity and inflated performance, while those that maintain separation between training and evaluation labels provide more reliable estimates of model capability. This perspective aligns with emerging data-centric approaches that emphasize the importance of data quality and labeling processes (Ng, 2021). This work reframes model evaluation itself as the object of study, rather than treating evaluation as a neutral measurement process.

## 7. Implications for Real-World Systems

The results of this study have direct implications for the design, evaluation, and deployment of machine learning systems in real-world information environments. In many operational settings, labels serve a dual role as both training data and evaluation targets. This dual use can obscure the true behavior of models and lead to overestimation of performance when evaluation is conducted using the same labeling source.

The findings suggest that practitioners should not rely solely on standard performance metrics derived from operational labels when assessing model quality. Instead, evaluation should incorporate independent or higher-authority label sources whenever possible. Even limited subsets of verified labels can provide valuable insight into the extent to which model performance reflects true predictive capability rather than alignment with a specific labeling system.

The divergence observed between ranking-based and classification-based metrics further highlights the need for careful interpretation of model outputs. In weakly supervised environments, models may retain useful signal for ranking or prioritization tasks, even when classification performance is unreliable. This suggests that deployment strategies should consider the appropriate use of model outputs, distinguishing between tasks that require precise classification and those that can tolerate approximate ranking.

The framework also underscores the importance of data governance in machine learning system design. Label quality is not a static property but is shaped by organizational processes, evolving standards, and human decision-making. As such, improving system performance requires not only advances in model architecture, but also attention to how labels are generated, maintained, and validated.

Finally, the concept of evaluation sovereignty provides a practical guideline for system design. Maintaining separation between training and evaluation label sources, where feasible, can reduce the risk of circular evaluation and improve the reliability of performance estimates. In cases where full separation is not possible, the multi-track evaluation framework offers a structured approach for identifying and quantifying potential bias.

## 8. Limitations

Several limitations of this study should be acknowledged. First, the gold evaluation subset used in Track R consists of 1,000 OSTI records annotated by a single reviewer under documented labeling guidelines derived from the official OSTI L1 label list and established normalization rules. While the gold subset is not intended as a universal ground truth, its purpose is to enforce independence from the operational labeling regime, enabling controlled evaluation of supervision effects. Because inter-annotator agreement was not computed, Track R results should be interpreted as a structured audit condition designed to enforce the supervisory split and reduce circular evaluation, rather than as independently adjudicated ground truth in the traditional sense. This constraint reflects practical limitations of the annotation process and is consistent with the exploratory role assigned to Track R within the multi-track framework.

Second, the empirical findings reported here are drawn from two domains — scientific metadata (OSTI) and biomedical literature (PubMed/MeSH). While these datasets provide complementary perspectives on operational and curated labeling environments, the generalizability of the framework and its findings to other information systems, taxonomies, and institutional labeling practices remains to be established. Future work should apply the multi-track framework to additional domains to assess the robustness of the evaluation sovereignty effect.

Third, the Track X condition, in which models trained on gold labels are evaluated against operational silver standards, was not empirically evaluated in this study. As discussed in Section 3, this inverted authority condition is theoretically significant because it predicts that models trained to a higher labeling standard may appear to underperform when evaluated against lower-authority labels. Empirical characterization of this effect represents a priority direction for future work and would complete the four-track framework introduced here.

Additionally, the OSTI Track F results should be interpreted as exploratory feasibility outcomes derived from operational labels, rather than authoritative evidence of true classification performance.

Furthermore, normalization procedures improved stability under operational (silver) supervision but did not resolve the underlying structural limitations associated with label authority and long-tail sparsity at Level-2.

Finally, model inputs were restricted to metadata fields (title and abstract) throughout all tracks. While this reflects realistic deployment constraints in large-scale repositories, it excludes full-text content that may carry additional signal relevant to fine-grained classification. The extent to which full-text inputs would alter the authority gap observed in Track R remains an open question.

## 9. Conclusion

This paper introduced evaluation sovereignty as a framework for analyzing how label authority and supervision regimes influence observed model performance. By explicitly distinguishing between operational ("silver") and independently verified ("gold") labels across multiple evaluation tracks, the proposed framework demonstrates that the authority and independence of label sources are primary determinants of what performance metrics actually measure.

Across all experiments, model performance varied substantially under different supervision regimes. Models that performed strongly under operational evaluation degraded significantly when assessed against independent labels, with the most pronounced effects observed in fine-grained classification. At Level-2, performance collapsed from approximately 0.54 to 0.03 Micro-F1, while ranking-based metrics remained non-trivial, revealing a divergence between latent model signal and classification validity. In contrast, models evaluated under curated supervision recovered strongly, indicating that the observed degradation reflects a governance gap rather than a limitation of model capability.

These findings demonstrate that evaluation validity is a system-level property shaped by the interaction between models, data, and label governance processes. Without explicit consideration of label authority, commonly reported metrics may reflect alignment with labeling processes rather than true predictive behavior. This has direct implications for model selection and deployment, as systems optimized against operational metrics may reinforce properties of the labeling process rather than the underlying phenomenon of interest.

The proposed multi-track framework provides a practical methodology for diagnosing evaluation bias and assessing reliability in real-world systems. By treating evaluation as an object of study, this work offers a foundation for improving measurement practices and supporting more trustworthy deployment of intelligent systems under weak supervision.

Future work will focus on extending this framework to additional domains and empirically characterizing the inverted authority condition (Track X), in which models trained on high-authority labels are evaluated against lower-authority standards. This direction will further clarify how label authority shapes evaluation outcomes and completes the framework introduced in this study.

**Data Availability**

The datasets used in this study are publicly available. OSTI records are accessible through the U.S. Department of Energy Office of Scientific and Technical Information (osti.gov), and PubMed data is available through the U.S. National Library of Medicine.

The manually verified gold subset used in Track R, along with experimental artifacts including data split definitions, label mappings, and configuration details, are available from the corresponding author upon reasonable request. These materials support reproducibility of the evaluation framework and experimental results.

**Declaration of Generative AI and AI-Assisted Technologies in the Writing Process**

During the preparation of this manuscript the author used Claude Sonnet 4.6 (Anthropic) to assist with structural review, formatting of results tables, and drafting of selected transitional passages. The author reviewed and edited all content and takes full responsibility for the publication.

**Declaration of Competing Interests**

The author declares that there are no known competing financial interests or personal relationships that could have appeared to influence the work reported in this paper.

**Funding**

This research did not receive any specific grant from funding agencies in the public, commercial, or not-for-profit sectors.